# Enhancing Building Semantics Preservation in AI Model Training with Large Language Model Encodings

Suhyung Jang[1], Ghang Lee[1,2], Jaekun Lee[1], and Hyunjun Lee[1]

[1]Building Informatics Group, Department of Architecture and Architectural Engineering, Yonsei University, Republic of Korea
[2]Institute for Advanced Studies, Technical University of Munich, Germany
rgb000@yonsei.ac.kr, glee@yonsei.ac.kr, jake102518@yonsei.ac.kr, hjlee97@yonsei.ac.kr

**Abstract**

Accurate representation of building semantics—encompassing both generic object types and specific subtypes—is essential for effective AI model training in the architecture, engineering, construction, and operation (AECO) industry. Conventional encoding methods (e.g., one-hot) often fail to convey the nuanced relationships among closely related subtypes, limiting AI's semantic comprehension. To address this limitation, this study proposes a novel training approach that employs large language model (LLM) embeddings (e.g., OpenAI GPT and Meta LLaMA) as encodings to preserve finer distinctions in building semantics. We evaluated the proposed method by training GraphSAGE models to classify 42 building object subtypes across five high-rise residential building information models (BIMs). Various embedding dimensions were tested, including original high-dimensional LLM embeddings (1,536, 3,072, or 4,096) and 1,024-dimensional compacted embeddings generated via the Matryoshka representation model. Experimental results demonstrated that LLM encodings outperformed the conventional one-hot baseline, with the "llama-3 (compacted)" embedding achieving a weighted average F1-score of 0.8766, compared to 0.8475 for one-hot encoding. The results underscore the promise of leveraging LLM-based encodings to enhance AI's ability to interpret complex, domain-specific building semantics. As the capabilities of LLMs and dimensionality reduction techniques continue to evolve, this approach holds considerable potential for broad application in semantic elaboration tasks throughout the AECO industry.

## 1 Introduction

The effective application of artificial intelligence (AI) technologies within the architecture, engineering, construction, and operation (AECO) industry hinges on the accurate representation of construction project information in machine-comprehendible formats. Extensive research has focused on utilizing diverse high-level data formats [1]—such as site photos for real-time site and worker condition monitoring [2,3], point clouds for tracking as-built construction progress [4–7], building information model (BIM) graphs for representing building topologies [8–10] and textual descriptions of project intents [11–13]—to represent information relevant to deliver the contextual information of the building projects. Representing and interpreting these digitally represented project information associated with planning, designing, constructing, and operating building projects (i.e., concepts that are significant for managing the projects), referred to as "building semantics" in this paper, are not only essential for AI systems to support reliable and informed decisions in construction projects, but also for their training.

To enable AI models to comprehend the high-level data format through supervised learning, this high-level information is labeled (partially or fully) and entered into AI models utilizing encoding methods. While these encodings play a critical role in differentiating the classes during training, prior studies often overlook the selection of encoding method, defaulting to conventional methods such as one-hot or label encodings. Although the advent of large language models (LLMs) and their generated embeddings has demonstrated the ability to capture domain-specific contextual nuances within the AECO fields [14,15], their utilization has primarily been confined to retrieving semantically similar information. The potential of leveraging these semantically rich LLM embeddings to provide task-relevant contextual nuances during AI model training remains largely unexplored.

This study proposes a novel AI model training method that employs LLM embeddings as encodings (i.e., "LLM encoding") to preserve finer distinctions between building semantics. To evaluate performance differences compared to conventional encoding methods, we conducted an experiment focusing on a building object subtype classification task using GraphSAGE models [16]. For 42 building object subtypes and 5 BIM models utilized in the practice by a major contractor in the Republic of Korea, GraphSAGE models were trained for the node classification task. The experiment was designed to contrast one-hot encoding with LLM encodings generated by various models, including OpenAI's 'text-embedding-3-small' and 'text-embedding-3-large' [17], as well as Meta's 'llama3' [18]. Additionally, we investigated diverse embedding dimensions, compacted using the Matryoshka representation model, to assess whether compacted LLM embedding still preserves semantic nuances during the training.

The remainder of this paper is organized as follows: Section 2 introduces the background and motivation for this study, including a review of previous research on semantic enrichment. Section 3 explains how LLM embeddings can be used as encodings to train AI models. Section 4 describes the experimental design used to validate the proposed method. Section 5 presents the experimental results and Section 6 discusses academic and practical implications, current limitations, and future studies. Finally, Section 7 concludes the paper.

## 2 Background

With the continually increasing efforts to employ AI in the AECO industry, there has been growing interest in training AI to understand the terms and concepts used in our field. In other words, there is a recognized need to teach computers the terminology and concepts we use to plan, design, construct, and operate construction projects. This study collectively refers to the information used in managing building projects as "building semantics."

The understanding of these building semantics by AI is largely achieved via supervised learning. In supervised learning, the objects of study—namely, the object, the concept (or class) in the target domain, and the encoding method used to distinguish and convey them—correspond to the components of the triangle of reference (i.e., the semantic triangle) [19], which are *referent*, *thought* (*reference*), and *symbol*. For example, in AI training aimed at classifying BIM object types, the wall objects in a BIM model (the *referent* in Figure 1) belong to the "Wall" category (*reference*). This might be represented to a computer as [1, 0, 0, 0, 0, 0], i.e., the first of six object types (*symbol*).

Previous research has shown that building semantics—that is, the diverse concepts or classes that exist in real or digital project representations—encompasses a wide range of meanings. For instance, to verify whether a construction worker on site is wearing the proper personal protective equipment (PPE), such as straps, hardhats, harnesses, and hooks, AI is used to recognize different types of PPE and check correct usage [2,3]. For studies focused on construction progress monitoring [4–7], AI that recognizes objects on an as-built construction site and tracks progress compared to the as-planned schedules. Other studies aim to predict object types or their subtypes in BIM models [8–10]. In short, there has been extensive research on enabling AI to interpret a broad range of building semantics.

Furthermore, the building objects used as examples of these concepts and classes, as well as the subjects of analysis, have been represented in ways that are suitable for AI comprehension. For instance, when monitoring the correct use of PPE, images of workers on site can be used; for as-built progress monitoring, point clouds can be adopted; and for object relationships in a BIM model, graph data can be employed. These representations help convey real or digitally depicted states to AI in forms appropriate for training.

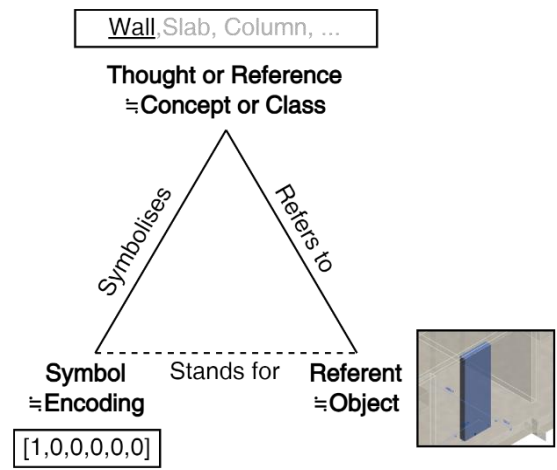

Figure 1. The triangle of reference (i.e., semantic triangle) in AI training for building semantics.

All these efforts show how to convey objects (the *referent*) to AI models, enabling them to interpret concepts (the *reference*) to understand building semantics. However, the types and effects of symbol—that is, the encodings used to help AI distinguish among concepts and classes—are comparatively underexplored. Although state-of-the-art large pretrained models can directly interpret concepts that are provided as images or text (i.e., can directly interpret referents), such models demand excessively high computational costs and datasets, making it difficult to train or fine-tune them for new purposes. Hence, it is more common to develop smaller models via supervised learning. In this context, conventional encoding methods that do not capture the relative similarities among classes and concepts (e.g., labels or one-hot encoding) have long been used without much question. For instance, one-hot encoding treats all categories as equidistant from one another, as illustrated in Figure 2A, where generic building object types ("Wall," "Column," "Foundation," "Beam," "Stair," and "Slab"), while ignores inherent similarities and differences between the categories.

This study proposes using semantically rich LLM embeddings (i.e., LLM encoding) as the encoding method for supervised training, thereby preserving building semantics. Unlike one-hot encoding, LLM embedding uses semantic knowledge learned from large training corpora to generate highly contextualized vector representations that can capture the relative similarities and differences among building semantics. Figures 2A and 2B illustrate how LLM embeddings represent semantic distances between generic object types (i.e., "Wall" and "Slab") and their subtypes using t-SNE visualization for OpenAI's 'text-embedding-3-large' (3,720 dimensions) and Meta's 'llama-3' (4,096 dimensions). In both Figures 2A and 2B, the embeddings form clearly distinct clusters of subtypes—one each for the generic types "Wall" and "Slab" (with a horizontal distinction in Figure 2B and a diagonal distinction in Figure 2C). In addition, both LLM embeddings recognize that parapet-related wall subtypes are more closely grouped together, compared to other wall subtypes (highlighted with gray backgrounds in Figures 2A and 2B). Comparing the x-axis and y-axis scales for OpenAI's GPT-based embedding in Figure 2B and Meta's LLaMA-based embedding in Figure 2C, it is evident that the two embeddings use different scales to distinguish each subtype (for example, $x \in [1,2]$, $y \in [-1.2,0]$ for 'text-embedding-3-large' in Figure 2B, and $x \in [1.2,2.6]$, $y \in [8,10]$ for 'text-embedding-3-large' in Figure 2C). Additionally, judging from the proportions of the ellipses around the parapet subtype clusters, we can see that the two LLM embeddings perceive the building semantics in subtly different ways.

Although it is reasonable to suspect that LLM embeddings might better symbolize the relative building semantics compared to one-hot encoding, this hypothesis remains underexplored. Building upon the premise that incorporating such building semantics into an AI model's encoding could enable more effective supervised learning, this study was conducted to propose a method that makes this approach feasible in practice.

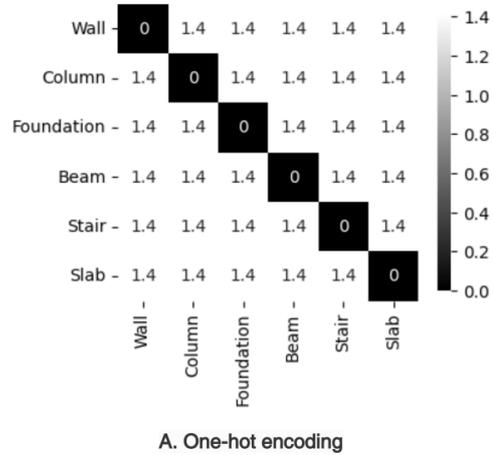

A. One-hot encoding

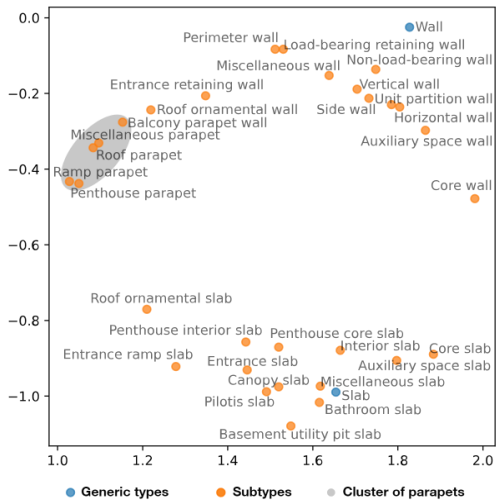

B. GPT embedding (text-embedding-3-large)

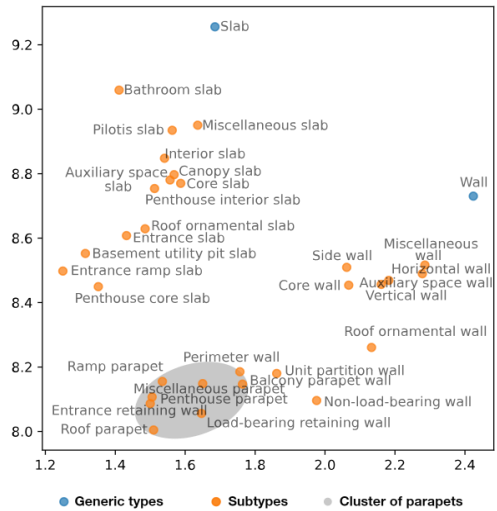

C. LLaMA embedding (llama-3)

Figure 2. Building semantics represented using diverse encoding types.

## 3 LLM Encoding and Matryoshka Representation Model

### 3.1 LLM Encoding

Typically, encodings are used to enable computers to distinguish between given classes. In neural network training, the model evaluates whether the output— derived from label encoding or one-hot encoding— matches the target class via a sigmoid function. If the output aligns with the target, the loss decreases, and training proceeds to minimize that loss. However, to utilize LLM embeddings as encodings (i.e., LLM encoding) in neural network training, the loss calculation method needs modification. LLM embeddings reside in high-dimensional spaces, capturing various semantic nuances. Applying a sigmoid function, which retains only the highest-value cell, can dilute these semantic features and hinder proper learning.

Instead of using a sigmoid at the final layer, LLM encoding sets the final layer's dimension to match that of the target LLM embedding. The loss is then computed based on the quantitative difference between the neural network's output embedding $e_p$ and the target label's LLM embedding $e_t$. One way to achieve this is by using cosine embedding loss, defined as equation (1):

$$L(e_p, e_t) = 1 - \frac{e_p \cdot e_t}{\|e_p\|\|e_t\|} \quad (1)$$

where · denotes the dot product and $\|\cdot\|$ denotes the Euclidean norm. This approach preserves semantic relationships in the embedding space, unlike one-hot encodings or sigmoid-based methods that force a single "correct" cell to dominate.

This training method is applicable to general tasks where each class carries semantic meaning. Once trained, the model can return the class closest to the predicted output by comparing the predicted embedding $e_p$ against each candidate embedding $e_t$ and selecting the one with the smallest loss.

### 3.2 Matryoshka Representation Model

LLMs generate embeddings in high-dimensional spaces, such as OpenAI's text-embedding-3-small (1,536 dimensions), text-embedding-3-large (3,072 dimensions), or Meta's LLaMA-3 (4,096 dimensions), to encode rich semantic information. However, leveraging these embeddings for real-world applications often requires dimensionality reduction to address computational inefficiencies. The Matryoshka representation model addresses this challenge by projecting high-dimensional embeddings into lower-dimensional spaces while preserving semantic features [20]. For implementation, the 'tomaarsen/mpnet-base-nli-matryoshka' model available on Hugging Face, was utilized.

## 4 Experiments

The experiment aims to evaluate whether utilizing LLM encoding improves the performance of AI models in enriching building semantics. Specifically, we assessed the AI model's performance on a subtype classification task of semantic elaboration, as depicted in Figure 3.

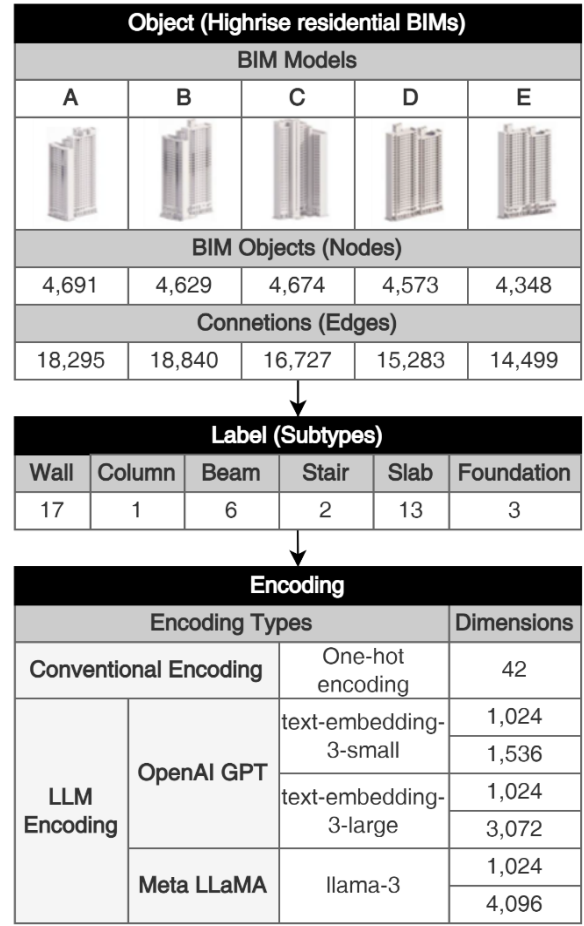

Figure 3. Experimental design.

For the experiment, five BIM models of high-rise residential projects, denoted A, B, C, D, and E, each containing over four thousand generic BIM objects, were utilized. Across these projects, we identified 42 subtypes managed by one of the major contractors in Republic of Korea (Table 1), ranging from "core wall" and "bathroom slab" to "transfer column" and "penthouse core slab from our prior study [10]. These projects were cross-validated by using each as a test dataset while the others served as training datasets. On these datasets, we trained GraphSAGE [21] models with three layers, each having a layer dimension of 1,024. Within this setup, we tested various encodings with varying dimension sizes using the Matryoshka Representation Model. The encodings included: (1) one-hot encoding, a conventional method; (2) 'text-embedding-3-small' with a 1,024-dimension compacted representation; (3) 'text-embedding-3-small' with its original 1,536 dimensions; (4) 'text-embedding-3-large' with a 1,024-dimension compacted representation; (5) 'text-embedding-3-large' with its original 3,072 dimensions; (6) 'llama-3' with a 1,024-dimension compacted representation; and (7) 'llama-3' with its original 4,096 dimensions.

Table 1. Subtypes utilized in the experiment.

| Code | Subtype | Code | Subtype |
|---|---|---|---|
| 0 | Core wall | 21 | Wall girder |
| 1 | Horizontal wall | 22 | Interior lintel |
| 2 | Vertical wall | 23 | Non-loadbearing lintel |
| 3 | Perimeter wall | 24 | Entrance stairs |
| 4 | Side wall | 25 | Interior stairs |
| 5 | Entrance retaining wall | 26 | Core slab |
| 6 | Loadbearing retaining wall | 27 | Entrance slab |
| 7 | Miscellaneous wall | 28 | Entrance ramp slab |
| 8 | Unit partition wall | 29 | Piloti slab |
| 9 | Non-loadbearing wall | 30 | Basement utility pit slab |
| 10 | Auxiliary space wall | 31 | Interior slab |
| 11 | Roof ornamental wall | 32 | Bathroom slab |
| 12 | Roof parapet | 33 | Miscellaneous slab |
| 13 | Penthouse parapet | 34 | Auxiliary space slab |
| 14 | Ramp parapet | 35 | Penthouse interior slab |
| 15 | Balcony parapet wall | 36 | Penthouse core slab |
| 16 | Miscellaneous parapet | 37 | Roof ornamental slab |
| 17 | Transfer column | 38 | Canopy slab |
| 18 | Core beam | 39 | Entrance strip foundation |
| 19 | Transfer beam | 40 | Mat foundation |
| 20 | Miscellaneous beam | 41 | Haunch |

To compare the impact of different encoding strategies on the model's ability to classify building object subtypes, the average F1-score was primarily utilized. In addition, statistical tests were conducted to determine the significance of the performance differences between encoding methods. The normality of F1-score differences was tested using the Shapiro-Wilk test with a significance level of α = 0.05. When the normality assumption was satisfied (p > 0.05), the paired t-test was employed to evaluate the statistical significance of the mean differences between paired observations. In cases where the normality assumption was violated (p ≤ 0.05), the Wilcoxon signed-rank test was utilized as a non-parametric alternative, with statistical significance set at α = 0.05.

## 5 Results

Table 2 presents the average F1-scores for classifying each subtype using both one-hot and LLM-based encodings. The encodings examined include text-embedding-3-small, text-embedding-3-large, and LLaMA-3, each tested in both compacted (1,024-dimensional) and original formats (1,536, 3,072, or 4,096 dimensions). Overall, subtypes with larger training and test sets demonstrated higher predictive performance.

Table 2. Weighted average F1-scores for different encoding types.

| Encoding type | Dimensions | Weighted average F1-score |
|---|---|---|
| **One-hot encoding** | 42 | 0.8475 |
| **text-embedding-3-small** | 1,024 | 0.8705 |
|  | 1,536 (original) | 0.8498 |
| **text-embedding-3-large** | 1,024 | 0.8655 |
|  | 3,072 (original) | 0.8529 |
| **llama-3** | 1,024 | 0.8766 |
|  | 4,096 (original) | 0.8714 |

With the original LLM embeddings, the weighted average F1-score increased sequentially from one-hot encoding (0.8475) to text-embedding-3-small (0.8498), text-embedding-3-large (0.8529), and finally LLaMA-3 (0.8714). However, when using the compacted 1,024-dimensional embeddings, the ranking of text-embedding-3-large and text-embedding-3-small reversed compared to their original counterparts, resulting in a slight overall increase in weighted average F1-scores to 0.8705, 0.8655, and 0.8766, respectively. These findings suggest that GraphSAGE, which outputs a 1,024-dimensional representation, may not adequately capture the semantic richness of higher-dimensional LLM embeddings (e.g., 1,536, 3,072, or 4,096 dimensions).

## 5.1 One-hot and LLM Encodings

In this subsection, we analyzed the significance of performance differences between one-hot and LLM encodings, including both original and compacted formats. The Shapiro-Wilk normality test (n=42) indicated that the normality assumption was violated in the F1-score differences for five out of six comparisons (p-values < 0.05), necessitating the use of the non-parametric Wilcoxon signed-rank test (Table 3). Among LLM embeddings, 'llama-3 (compacted)' exhibited a distinctly low p-value (0.000007), indicating a more skewed fold-by-fold performance distribution that may arise from uneven semantic feature retention after the largest dimensional compression (4,096 to 1,024).

Table 3. p-values of the Shapiro-Wilk test for identifying the normality of the difference in performance between one-hot and LLM encodings.

| Shapiro-Wilk test | One-hot encoding |
|---|---|
| text-embedding-3-small (compacted) | 0.065730 |
| text-embedding-3-small (original) | 0.006618 |
| text-embedding-3-large (compacted) | 0.013222 |
| text-embedding-3-large (original) | 0.011914 |
| llama-3 (compacted) | 0.000007 |
| llama-3 (original) | 0.049937 |

The results of the paired t-test and Wilcoxon signed-rank test (n=42) are presented in Table 4. Although the performance of the model slightly improved when LLM encodings were utilized, statistical analysis revealed that no significant differences were found, except for 'text-embedding-3-large (compacted)' (p=0.006596). This suggests that compressing text-embedding-3-large to 1,024 dimensions appears to preserve key semantic cues while removing noise, thus yielding more consistent performance gains and a statistically significant improvement over one-hot encoding.

Table 4. p-values of paired t-test and Wilcoxon signed-rank test for identifying the significance of the difference in performance between one-hot and LLM encodings.

| Paired t-test and Wilcoxon signed-rank test | One-hot encoding |
|---|---|
| text-embedding-3-small (compacted) | 0.285281 (z=1.06852) |
| text-embedding-3-small (original) | 0.873084 (t = -0.16074) |
| text-embedding-3-large (compacted) | 0.006569 (z=2.71792) |
| text-embedding-3-large (original) | 0.163854 (z=1.39222) |
| llama-3 (compacted) | 0.714148 (z= 0.36629) |
| llama-3 (original) | 0.219283 (z= 1.22843) |

## 5.2 Between LLM Encodings

In this subsection, we analyzed the statistical significance between different LLM encodings. Similar to the one-hot comparisons, the Shapiro-Wilk test (n=42) identified violations of normality assumptions in all cases, leading us to employ the Wilcoxon signed-rank test (n=42) for all comparisons. For original LLM embeddings, no significant differences were observed despite slight performance increases (Table 5).

Table 5. p-values of Wilcoxon signed-rank test for identifying the significance of the difference in performance between original LLM encodings.

| Wilcoxon signed-rank test | text-embedding-3-large (original) | llama-3 (original) |
|---|---|---|
| text-embedding-3-small (original) | 0.068390 (z = 1.82243) | 0.529727 (z = 0.62842) |
| text-embedding-3-large (original) | - | 0.751786 (z = 0.31628) |

Conversely, for compacted LLM encodings, significant differences were observed (Table 6). This demonstrates that compacted LLM embeddings effectively leverage the semantic richness provided by large datasets and advanced architectures. For example, 'llama-3 (compacted)' benefits from its extensive training corpus of over 1.4 trillion tokens and 65 billion model parameters, compared to the smaller 'text-embedding-3-large (compacted)'.

Table 6. p-values of Wilcoxon signed-rank test for identifying the significance of the difference in performance between compacted LLM encodings.

| Wilcoxon signed-rank test | text-embedding-3-large (compacted) | llama-3 (compacted) |
|---|---|---|
| text-embedding-3-small (compacted) | 0.010346 (z = 2.56403) | 0.011765 (z = 2.51910) |
| text-embedding-3-large (compacted) | - | 0.000583 (z = 3.43962) |

These results suggest that as LLM technology continues to evolve, the utility of LLM encodings for AI-based training will further improve. A notable observation is the consistent outperformance of compacted LLM encodings compared to their original counterparts, highlighting the effectiveness of dimensionality reduction techniques like the Matryoshka representation model in preserving essential semantic information. Additionally, the findings imply that increasing the size of AI models may be necessary to fully harness the potential of semantically rich LLM encodings. Future research should explore the optimal balance between model size and computational efficiency to maximize performance in semantic elaboration tasks.

## 6 Discussion

### 6.1 Academic and Practical Implications

This study underscores the importance of integrating semantically rich encodings into AI models to enhance the understanding of building semantics. From an academic perspective, the approach broadens current research by demonstrating that LLM embeddings—when compacted appropriately—can outperform conventional one-hot methods, thereby offering a more nuanced way to represent complex object subtypes. Practitioners, on the other hand, gain a practical encoding method that leverages widely available pretrained LLMs for improved decision support. By preserving finer distinctions among building semantics, AI models can more accurately comprehend real-world project information, ultimately helping industry professionals make more context-aware informed decisions supported by AI models.

### 6.2 Limitations

Despite its promising outcomes, this study has several limitations that call for cautious interpretation. First, the experiments limitedly covered a subtype classification task involving 42 subtypes across five high-rise residential BIM models. These may not comprehensively reflect the full range of building types or the myriad semantic tasks encountered in real-world practice. Additional work is needed to validate the proposed method on more diverse building typologies (e.g., hospitals, commercial complexes, infrastructure) and under different project conditions (e.g., varying project phases, levels of detail). Second, while the focus here was on subtype classification using a graph neural network, building semantics encompasses a broad range of tasks—such as construction scheduling, cost estimation, and design intent capture—that might benefit differently from LLM encodings. Lastly, the reliance on existing LLMs means that domain-specific terms not extensively represented in their training corpora may still suffer from suboptimal embeddings, highlighting the need for either fine-tuning or training domain-specific LLMs on specialized AECO datasets.

### 6.3 Future Directions

Building on the insights gained in this study, several avenues for future work emerge. First, exploring alternative loss functions—beyond cosine similarity—may further refine model training for specific semantic tasks. Second, investigating different dimensionality-reduction techniques could help preserve even more nuanced contextual information while mitigating the computational overhead of high-dimensional embeddings. Third, training or fine-tuning LLMs on construction-centric corpora, such as project specifications or technical standards, may yield embeddings that better capture domain-specific language and nuances. Finally, a more holistic approach that combines text-based encodings with additional data modalities—such as 2D/3D images, point clouds, or sensor data—could offer a deeper, multimodal understanding of building semantics. By expanding the scope of tasks and data types, future research can further validate the robustness and applicability of LLM encodings in increasingly complex AECO environments.

## 7 Conclusion

This paper tackled the challenge of preserving and enriching building semantics in AI models by introducing a training method that employs LLM embeddings as encodings. Through extensive experiments on BIM object subtype classification, we demonstrated that compacted LLM encodings, such as "llama-3 (compacted)," outperform conventional one-hot encoding, achieving higher weighted average F1-scores (0.8766 over 0.8575) and showing stronger potential to represent subtle semantic distinctions. Although statistical significance was not uniformly observed for all embeddings, the results highlight the growing promise of leveraging advanced language models to capture the domain-specific context vital to the AECO industry.

By integrating LLM-based encodings into class labeling for supervised AI model training, this study contributes a viable solution for domain practitioners and researchers seeking to improve model accuracy and semantic fidelity. As LLM technology continues to advance—and as construction datasets become richer—opportunities to refine these methods will likely multiply. Ultimately, the framework presented here offers a stepping stone toward more intelligent, context-aware AI systems capable of handling the increasing complexity and specialization inherent in modern building projects.

## Acknowledgements

This work is supported in 2025 by the Korea Agency for Infrastructure Technology Advancement (KAIA)

grant funded by the Ministry of Land, Infrastructure and Transport (Grant RS-2021-KA163269 and RS-2024-00407028) and the Hans Fischer Senior Fellowship program at the Technical University of Munich - Institute for Advanced Studies (TUM-IAS) in Germany.

## References

[1] S. Jang, H. Roh, G. Lee, Generative AI in architectural design: Application, data, and evaluation methods, Automation in Construction 174 (2025) 106174. https://doi.org/10.1016/j.autcon.2025.106174.

[2] D. Gil, G. Lee, Zero-shot monitoring of construction workers' personal protective equipment based on image captioning, *Automation in Construction*, 164:105470, 2024.

[3] W.-C. Chern, J. Hyeon, T.V. Nguyen, V.K. Asari, H. Kim, Context-aware safety assessment system for far-field monitoring, *Automation in Construction* 149:104779, 2023.

[4] J. Ma, J. Jung, F. Leite, Deep Learning–Based Scan-to-BIM Automation and Object Scope Expansion Using a Low-Cost 3D Scan Data, *Journal of Computing in Civil Engineering*, 38:04024040, 2024.

[5] J. Kim, J. Kim, N. Koo, H. Kim, Automating scaffold safety inspections using semantic analysis of 3D point clouds, *Automation in Construction*, 166:105603, 2024.

[6] R. Sacks, L. Ma, R. Yosef, A. Borrmann, S. Daum, U. Kattel, Semantic Enrichment for Building Information Modeling: Procedure for Compiling Inference Rules and Operators for Complex Geometry, *Journal of Computing in Civil Engineering*, 31:04017062, 2017.

[7] J.W. Ma, T. Czerniawski, F. Leite, Semantic segmentation of point clouds of building interiors with deep learning: Augmenting training datasets with synthetic BIM-based point clouds, *Automation in Construction*, 113:103144, 2020.

[8] Z. Wang, R. Sacks, B. Ouyang, H. Ying, A. Borrmann, A Framework for Generic Semantic Enrichment of BIM Models, *Journal of Computing in Civil Engineering*, 38:04023038, 2024.

[9] G. Austern, T. Bloch, Y. Abulafia, Incorporating Context into BIM-Derived Data—Leveraging Graph Neural Networks for Building Element Classification, *Buildings*, 14:527, 2024.

[10] S. Jang, G. Lee, M. Park, J. Lee, S. Suh, B. Koo, Semantic elaboration of low-LOD BIMs: Inferring functional requirements using graph neural networks, *Advanced Engineering Informatics*, 64: 103100, 2025.

[11] S. Jang, G. Lee, Interactive Design by Integrating a Large Pre-Trained Language Model and Building Information Modeling, In *Proceedings of Computing in Civil Engineering 2023*, pages 291-299, Corvallis, Oregon, United States, 2023.

[12] S. Jang, G. Lee, J. Oh, J. Lee, B. Koo, Automated detailing of exterior walls using NADIA: Natural-language-based architectural detailing through interaction with AI, *Advanced Engineering Informatics*, 61:102532, 2024.

[13] M. Uhm, J. Kim, S. Ahn, H. Jeong, H. Kim, Effectiveness of retrieval augmented generation-based large language models for generating construction safety information, *Automation in Construction*, 170:105926, 2025.

[14] Y. Gao, G. Xiong, H. Li, J. Richards, Exploring bridge maintenance knowledge graph by leveraging GraphSAGE and text encoding, *Automation in Construction*, 166:105634, 2024.

[15] J. Lee, S. Ahn, D. Kim, D. Kim, Performance comparison of retrieval-augmented generation and fine-tuned large language models for construction safety management knowledge retrieval, *Automation in Construction*, 168:105846, 2024.

[16] G. Lee, S. Jang, S. Suh, M. Park, H. Roh, L. Kim, Three approaches for building information modeling library transplant, in: Seoul, Korea, 2023.

[17] A. Neelakantan, T. Xu, R. Puri, A. Radford, J.M. Han, J. Tworek, Q. Yuan, N. Tezak, J.W. Kim, C. Hallacy, J. Heidecke, P. Shyam, B. Power, T.E. Nekoul, G. Sastry, G. Krueger, D. Schnurr, F.P. Such, K. Hsu, M. Thompson, T. Khan, T. Sherbakov, J. Jang, P. Welinder, L. Weng, Text and Code Embeddings by Contrastive Pre-Training, On-line: https://doi.org/10.48550/arXiv.2201.10005, Accessed: 12/23/2024.

[18] H. Touvron, T. Lavril, G. Izacard, X. Martinet, M.-A. Lachaux, T. Lacroix, B. Rozière, N. Goyal, E. Hambro, F. Azhar, A. Rodriguez, A. Joulin, E. Grave, G. Lample, LLaMA: Open and Efficient Foundation Language Models, On-line: https://doi.org/10.48550/arXiv.2302.13971, Accessed: 12/23/2024.

[19] C.K. Ogden, I.A. Richards, *The Meaning of Meaning: A Study of the Influence of Language upon Thought and of the Science of Symbolism*, Harcourt, Brace, 1927.

[20] A. Kusupati, G. Bhatt, A. Rege, M. Wallingford, A. Sinha, V. Ramanujan, W. Howard-Snyder, K. Chen, S. Kakade, P. Jain, Matryoshka representation learning, *Advances in Neural Information Processing Systems*, 35:30233–30249, 2022.

[21] W. Hamilton, Z. Ying, J. Leskovec, Inductive representation learning on large graphs, Advances in Neural Information Processing Systems, 30, 2017.